# Guided Policy Exploration for Markov Decision Processes using an Uncertainty-Based Value-of-Information Criterion

Isaac J. Sledge, *Student Member, IEEE*, Matthew S. Emigh, *Member, IEEE*, and José C. Príncipe, *Life Fellow, IEEE*

**Abstract**—Reinforcement learning in environments with many action-state pairs is challenging. At issue is the number of episodes needed to thoroughly search the policy space. Most conventional heuristics address this search problem in a stochastic manner. This can leave large portions of the policy space unvisited during the early training stages. In this paper, we propose an uncertainty-based, information-theoretic approach for performing guided stochastic searches that more effectively cover the policy space. Our approach is based on the value of information, a criterion that provides the optimal trade-off between expected costs and the granularity of the search process. The value of information yields a stochastic routine for choosing actions during learning that can explore the policy space in a coarse to fine manner. We augment this criterion with a state-transition uncertainty factor, which guides the search process into previously unexplored regions of the policy space.

We evaluate the performance of uncertainty-based value-of-information policies on aspects of the games Centipede and Crossy Road. Our results indicate that the uncertainty-based value-of-information criterion yields better-performing policies in fewer episodes than stochastic-based exploration strategies. We show that the training rate for our approach can be further improved by using the policy cross-entropy to guide our criterion's hyperparameter selection.

**Index Terms**—Value of information, exploration, information-theoretic exploration, uncertainty-based exploration, reinforcement learning, information theory

## 1 Introduction

The problem of optimal decision making under uncertainty is crucial for intelligent agents. Reinforcement learning [1] addresses this problem by proposing that agents should maximize an expected long-term return provided by the environment. This provides the basis for trial-and-error-based learning of an action-selection policy.

There are a variety of reinforcement learning techniques that can be applied for uncovering policies. An issue faced by these techniques is how to uncover high-performing policies. In some instances, it may be necessary for an agent to try seemingly non-cost-optimal actions at different stages in learning process. This is referred to as exploration, as it causes the search process to visit potentially novel regions of the policy space. In other instances, it may be prudent for an agent to capitalize on already-established, cost-effective solutions. This is known as exploitation, as the agent is leveraging the current policy to choose an appropriate action at a given state.

A balance between exploration and exploitation is needed for many problems [2]. Without an ability to adapt, agents may complete tasks well at one phase of the learning process and not in a later one. This is likely to occur when the agents operate in stochastic environments whose dynamics change. Likewise, without an ability to exploit, the agent may continuously elect to carry out actions that lead to poor long-term expected costs.

There have been several methods developed that attempt to best trade-off between these two competing processes [3]. A defining trait of many of the conventionally used methods is that they are stochastic. That is, they choose which actions to take in either a random or a cost-driven, weighted-random fashion. This, however, can pose some practical difficulties. They may fail to visit large portions of the policy space, where the environment dynamics are poorly understood, when training over a finite number of episodes. They may also frequently revisit the same regions in the policy space without meaningfully updating the policy. Sub-par policies can hence be returned if the learning process is terminated early.

Isaac J. Sledge is with the Department of Electrical and Computer Engineering, University of Florida, Gainesville, FL 32611, USA (email: isledge@ufl.edu). He is also with the Computational NeuroEngineering Laboratory (CNEL) at the University of Florida.

Matthew S. Emigh was with the Department of Electrical and Computer Engineering, University of Florida, Gainesville, FL 32611, USA (email: matt@cnel.ufl.edu). He was also with the Computational NeuroEngineering Laboratory (CNEL) at the University of Florida. He is currently with the United States Naval Surface Warfare Center PCD, Panama City, FL 32407, USA.

José C. Príncipe is the Don D. and Ruth S. Eckis Chair and Distinguished Professor with both the Department of Electrical and Computer Engineering and the Department of Biomedical Engineering, University of Florida, Gainesville, FL 32611, USA (email: principe@cnel.ufl.edu). He is the director of the Computational NeuroEngineering Laboratory (CNEL) at the University of Florida.

The work of the authors was funded via grant N00014-15-1-2103 from the US Office of Naval Research. The first author was additionally funded by a University of Florida Research Fellowship, a Robert C. Pittman Research Fellowship, and an ASEE Naval Research Enterprise Fellowship.

For exploration to be effective, the search process should systematically target regions of the policy space where the environment dynamics are poorly understood. In doing so, the agent can quickly adapt its behaviors to the environment dynamics, leading to high-performing policies early during training. In this paper, we propose a hybrid model-based and model-free framework for addressing this desire, which is based on the value of information [4, 5].

The value of information implements an information-theoretic [6] optimization criterion. It describes the maximum benefit that can be gained by a given quantity of information in order to minimize average losses. When applied to Markov decision processes, the criterion provides an optimal trade-off between policy cost and the granularity of the search process [7]. The granularity is dictated by the number of bits needed to encode the policy. When the number of bits is low, the amount of information that the states provide about the actions is artificially limited. The policy space is searched coarsely and the exploitation of previously learned behaviors dominates. This is due to a quantization of the action-state space: many states will be grouped together and visiting them will elicit the same action response. When the number of bits is high, the amount of information carried about the actions by the states is increased. The policy space is thoroughly surveyed and the exploration of new actions dominates. There is little to no quantization of the action-state space in this case.

Optimizing the value of information provides a weighted-random action-selection scheme for reinforcement learning [7, 8]. Our contribution in this paper is a novel extension of the value of information that causes the exploration process to systematically target regions of the policy space where the dynamics are poorly understood. We do this by adding a model-based component which captures the future-state predictive capability of a model that is constructed during learning [9]. Attempting to maximize the prediction accuracy of this model while also minimizing expected agent costs directs the search process to regions where there is high prediction uncertainty. Once such regions are investigated and the model accuracy improves, the search process can move to other high-uncertainty areas where the prediction rate is lower and repeat the process. The entire policy space is efficiently explored in a semi-uniform fashion early during training, which quickly leads to high-performing policies.

To evaluate this uncertainty-based value-of-information criterion, we consider two games. The first is a simplified variant of the arcade game Centipede. We highlight how our uncertainty-based criterion explores the limited state-action space more effectively than purely stochastic search approaches. We also investigate the influence of our criterion's hyperparameters on the agent performance. The second game that we consider is Crossy Road, an endless version of the classic arcade game Frogger. We consider this game to demonstrate that our criterion scales well to much more complicated state-action spaces. It is also used to again showcase that far fewer episodes are required to specialize to this environment than purely random exploration styles.

Our simulations also provide insights into facilitating efficient learning. Reinforcement learning with conventional exploration heuristics often relies on an exploration-intensive search at the beginning of the learning process. This search eventually gives way to one that is exploitation-intensive, so as to retain meaningful agent behaviors. When using the uncertainty-based value of information, the situation should be reversed. Investigators should initially favor an exploitation-intensive search. This leads to a coarse quantization of the space, which can be searched quickly in a semi-uniform fashion. The quantization amount can then be iteratively refined to promote the targeted exploration of regions with low-cost policies.

We propose to use an information-theoretic quantity, the cross-entropy between consecutive policies, to help decide when to adjust the quantization level. Cross-entropy is an appealing quantity for this purpose. It provides a bounded measure of change between pairs of probabilistic policies. It hence roughly describes the rate of learning. Cross-entropy magnitudes near one correspond to many entries in a tabular policy being changed. Values of cross-entropy near zero imply that few changes are being made to a tabular policy. Our use of this quantity for automatically adjusting hyparparameter values is another novel contribution of this paper.

The remainder of this paper is organized as follows. We begin, in section 2, with a survey of uncertainty-based exploration for reinforcement learning. In section 3.1, we give a review of the value of information from a Markov-decision-processes perspective. Our uncertainty-based extension of the criterion, which is the focus of this paper, is presented in section 3.2. We combine this criterion with $Q$-learning in section 3.2.1. We also discuss practical aspects of it in this section.

Our simulation results for Centipede and Crossy Road are given in sections 4 and 5, respectively. We begin by covering our experimental protocols. The games' reward structures are given in sections 4.1.1 and 5.1.1. In sections 4.1.2 and 5.1.2, we discuss what features we use for reinforcement learning. In sections 4.2 and 5.2, we present our simulation results. In section 4.2.1, we cover the quantitative effects of the hyperparameters on the learning process and the qualitative effects on agent behaviors. Comparisons against uncertainty-based exploration approaches are provided in section 4.2.2. In section 5.2.1, we cover the qualitative and quantitative improvement in the agent's behaviors for fixed hyperparameter values. In section 5.2.2, we analyze the policy performance improvements when using our cross-entropy heuristic to automatically adjust hyperparameter values. Performance comparisons of our policies against those from stochastic exploration heuristics, like epsilon-greedy and soft-max-based search, are provided in section 5.2.3.

## 2 Literature Comparisons

Many exploration schemes have been advanced over the years. Out of these, the most related to our work are those that implement uncertainty-driven search. In uncertainty-based exploration, if a state has not been visited sufficiently for the agent to be familiar with it, then that state will have high uncertainty and an agent will be driven to it. The agent will attempt to experience all possible transitions from that state. This ensures that an agent will thoroughly investigate new areas of the action-state space.

One of the earliest uncertainty-based exploration strategies was given in [10]. In [10], Whitehead proposed a counter-based rule that weights each action in proportion to how many steps have been taken since that action has last been used. Actions are selected at random according to this weighting. He showed that, for finitely sized environments, the goal state is always found in time that is polynomial in the size of the state space when choosing actions in this manner. This assumes either access to the transition probabilities or that such transitions can be reliably estimated, neither of which is easily feasible for some environments. Similar results were obtained for a related approach developed by Sutton [11].

In [12, 13], Kearns, Singh, and Koller presented the uncertainty-based, explicit-explore-or-exploit algorithm and provided sample complexity results for it. Their approach was similar to Whitehead's [10], in the sense that they maintain a list of how many times each state has been visited. If a state has been visited a sufficient number of times, then it is added to a so-called known-state list and either exploitation of the current policy or exploration is performed for that state. If the agent transitions to a state that is not on the known-state list, then the action chosen the fewest number of times at that state is taken. When following such a procedure, the authors guaranteed polynomial convergence time to the goal state. This contrasts with the asymptotic results for many reinforcement learning methodologies [14, 15], which do not bound the number of actions.

Brafman and Tennenholtz proposed one of the first practical implementations of the explicit explore or exploit algorithm [16, 17]. A downside of this implementation is that it requires visiting every state exhaustively, which is impractical for many problems. Strehl et al. [18–20] performed extensive theoretical analyses of many uncertainty-based exploration approaches, including the explicit-explore-or-exploit algorithm. They provided sample-complexity results both for reinforcement learning methods that learn a model and methods that learn directly a value function. Saul and Singh furnished related results for special classes of Markov decision process problems designed to highlight a particular exploration-exploitation trade-off [21]. Fiechter [22, 23] did the same for a learning protocol that can return to a set of starting states at arbitrary times. Lastly, Schapire and Warmuth [24] formulated non-asymptotic bounds for uncontrolled Markov processes.

In [25], Dearden, Friedman, and Andrew presented a model-based reinforcement learning method in which a parametric distribution over the action-value function is estimated via Bayesian inference. The distribution captures uncertainty in the estimate of the action-value function. This criterion explores an action only if enough uncertainty remains such that it is possibly the optimal action. After sufficient exploration, the variances in the action-value function estimates are small enough that it is unlikely any action other than the highest rated action is optimal. While this method provides a measure of uncertainty that is local in the state-action space, it is only practical for small, discrete reinforcement learning problems. We refer to [26, 27] for related notions and to [28–33] for more details about Bayesian reinforcement learning.

Emigh, Kriminger, and Príncipe proposed a model-based divergence-to-go framework for exploration in [9]. Their idea was to employ a divergence to measure recursion similarity and hence quantify the transition uncertainty associated with each state-action pair. This divergence distribution was constructed in an iterative fashion in a manner similar to the value functions in temporal-difference learning. However, unlike a value function, the divergence-to-go is not a stationary quantity that is estimated for each state. Instead, as the transition models improve, the divergence values will decrease to zero. Choosing actions based on the maximum divergence-to-go has the effect of maximizing exploration, in the sense of either visiting the states or experiencing the state transitions, both of which effectively learn the state transition model. Just as with choosing the action with the largest value function magnitude [34], the divergence-to-go accounts for future uncertainty as well.

Our approach in this paper is most related to that of Emigh et al. [9]. We favor a model-based approach for uncertainty that is based on expected divergence. This model attempts to predict the next-state transitions. Minimizing prediction uncertainty causes the policy iterates to jump to regions of the policy space whose dynamics are not well understood by that point in the learning process. Unlike the approach in [9], we combine this model with the model-free value-of-information criterion. The value of information can be used to quantize the state space, which simplifies the search for high-performing policies. States are grouped and assigned a single action. Using uncertainty-based searches in in this quantized policy space facilitates efficient, semi-uniform exploration. Our empirical results corroborate this claim: we obtain lower-cost policies in fewer episodes than is required by many reinforcement learning heuristics. We are also able to apply our approach to relatively high-dimensional action-state spaces; in contrast, many existing uncertainty-based exploration schemes are not tractable past a few dimensions [12, 13, 16, 17].

## 3 Methodology

We seek a means to determine when it is appropriate to choose actions that deviate from the policy and when it is not. Appropriateness is reflected in the expected costs that the agent will accrue. We also seek a way to specify how cost-sensitive the action-selection process should be.

In our previous work, we showed that these desires can be realized by leveraging information that the states carry about the actions [35]. This idea of utilizing information in decision-making was developed into a rigorous theory by Stratonovich [4, 5]. This theory took the form of a value of information criterion, which we previously investigated from a reinforcement-learning perspective in [7, 8]. The criterion describes the maximum benefit obtained from a piece of information for reducing average costs.

In [8], we analyzed the exploration optimality of the value of information for single-state, multi-action Markov decision processes. We proved that value-of-information-based policies was an optimal search process, provided that the number of bits was systematically adjusted during learning. We also empirically demonstrated that our approach would consistently outperform others specifically formulated for this class of problems.

In [7], we formulated the value of information to explore multi-state, multi-action Markov decision processes. We analyzed the connections between the number of bits and emergent agent behaviors. We demonstrated that certain agent behaviors would only arise due to a sufficient level of action-selection risk, as quantified by von-Neumann-Morgenstern utility theory.

In this section, we build upon our work in [7, 8]. We review the stochastic value-of-information criterion [7] in section 3.1. In section 3.2, we show how this criterion can be augmented with an uncertainty-based, transition-model component. This novel extension allows value-of-information-based search to systematically target regions of the policy space where the transition dynamics are poorly understood. Thoroughly investigating such areas leads to the quick formation of high-performing policies. Purely stochastic, model-free searches, in comparison, may need many times more learning episodes to adequately visit these areas, as they have no mechanism to directly assess the environment dynamics. The original value of information criterion also suffers from this issue, so this modification is crucial for obtaining good practical performance without relying on heuristics like experience replay.

In section 3.2.1, we build upon our findings from [8]. We provide a novel information-theoretic heuristic for automatically tuning the number of bits during learning. Using this heuristic increases the policy improvement rate compared to fixed-parameter searches.

### 3.1 Value of Information: Random Exploration Case

Consider a composite system defined by a measurable state space $\mathcal{S}$ and a measurable action space $\mathcal{A}$. We assume that the state $s_t \in \mathcal{S}$ visited at discrete time $t$ is a random variable described by some distribution. After the observation of the value $s_t$, one chooses an optimal estimator $a_t \in \mathcal{A}$ at that time instant which minimizes the conditional expected penalty, $\inf_{a_t \in \mathcal{A}} \mathbb{E}[Q(s_t, a_t)|s_t]$. Averaging the conditional expected penalties yields the total expected penalty, $\mathbb{E}[\inf_{a_t \in \mathcal{A}} \mathbb{E}[Q(s_t, a_t)|s_t]]$. Here $Q : \mathcal{S} \times \mathcal{A} \to \mathcal{R}$ is the penalty term, such as an action-value function associated with agent costs.

There are two extreme cases to consider when finding an action-selection policy that solves the total expected penalty criterion. The first case is when no information about the value of the random variable $s_t \in \mathcal{S}$ is available. That is, states carry no information about the actions that should be selected. There is only one way to choose the optimal estimator $a_t \in \mathcal{A}$ when this occurs: minimize the average penalties $\mathbb{E}[\inf_{a_t \in \mathcal{A}} \mathbb{E}[Q(s_t, a_t)|s_t]] = \inf_{a_t \in \mathcal{A}} \mathbb{E}[Q(s_t, a_t)]$. The action-selection policy is hence a uniform distribution, there is complete uncertainty over what action should be chosen. On the other hand, if the states carry total information about the actions, then $\mathbb{E}[\inf_{a_t \in \mathcal{A}} \mathbb{E}[Q(s_t, a_t)|s_t]] = \mathbb{E}[\inf_{a_t \in \mathcal{A}} Q(s_t, a_t)]$. In this instance, the action-selection policy is a delta function, as there is no uncertainty about the optimal estimator.

For many problems, only partial information is available. There is a challenge of how exactly to best use it. One way of doing this is through the value of information [4, 5], which connects both the complete-information and no-information cases with a smooth, non-linear transition that relates obtainable costs and information. That is, it provides an optimal conversion between varying degrees of partial information and expected obtainable costs.

The value of information combines Shannon's information theory with the concept of average losses or risk, which characterizes the quality of decisions being made. This criterion can be defined as the maximum benefit that can be gained by a given quantity of information in order to minimize average losses. As such it is an independent branch of information theory that can be used in many practical applications.

For Markov-decision-process-based reinforcement learning, the criterion can be utilized to choose optimal-cost actions for each state. This is done by minimizing the difference in average agent costs for the no-information case with the total expected agent costs for the partial-information case,

$$\underbrace{\inf_{a_t \in \mathcal{A}} \left( \sum_{s_t \in \mathcal{S}} p(s_t) Q(s_t, a_t) \right)}_{\inf_{a_t \in \mathcal{A}} \mathbb{E}\left[ Q(s_t, a_t) \right]} - \underbrace{\inf_{\pi_{s_{t+1}}(a_t, s_t)} \left( \sum_{s_t \in \mathcal{S}} \sum_{a_t \in \mathcal{A}} p(s_t) \pi_{s_{t+1}}(a_t, s_t) Q(s_t, \pi^*_{a'_t}(s_t)) \right)}_{\inf_{\pi_{s_{t+1}}(a_t, s_t)} \mathbb{E}\left[ \inf_{a'_t \in \mathcal{A}} \mathbb{E}\left[ Q(s_t, a'_t) \middle| s_t \right] \right]}. \tag{3.1}$$

Here, we use $\pi_{s_{t+1}}(a_t, s_t)$ to represent a stochastic action-selection policy. It is the probability of choosing action $a_t \in \mathcal{A}$ when in state $s_t \in \mathcal{S}$ and then transitioning to state $s_{t+1} \in \mathcal{S}$. The term $\pi^*_{a'_t}(s_t)$ represents the best action $a'_t \in \mathcal{A}$ chosen when using the optimal policy.

For the second term in (3.1), we have that the conditional probabilities representing the policy are subject to an information constraint, which we take to be Shannon mutual information

$$\pi_{s_{t+1}}(a_t, s_t) \text{ such that} : \underbrace{\sum_{s_t \in \mathcal{S}} p(s_t) \sum_{a_t \in \mathcal{A}} \pi_{s_{t+1}}(a_t, s_t) \log\left( \frac{\pi_{s_{t+1}}(a_t, s_t)}{p(a_t)} \right)}_{\mathbb{E}\left[ D_{\mathrm{KL}}(\pi_{s_{t+1}}(a_t, s_t) \| p(a_t)) \right]} \leq \varphi_{\text{inf}},\ \varphi_{\text{inf}} > 0. \tag{3.2}$$

This constraint is parameterized by a positive, user-selectable value $\varphi_{\text{inf}}$. This value artificially limits how much information the states should carry about what actions should be taken.

The criterion defined by (3.1) and (3.2) define a difference in expected costs. They can be described as follows:

**First Term: No-Information Returns.** The first term $\inf_{a_t \in \mathcal{A}} \mathbb{E}[Q(s_t, a_t)]$ captures the possible returns for a policy in which no information about the actions can be inferred from the states. This is used to establish the baseline agent performance, as it is the worse possible cost that can be obtained by the agent. If the states are not informative, then the optimal action is based solely on the state random variable distribution. If, however, the states are informative, then the returns for the simplest policy will be offset by a second term.

**Second Term: Informative Returns.** The second term is based on the expected return using a modified action-value function, $\inf_{\pi_{s_{t+1}}(a_t, s_t)} \mathbb{E}[\inf_{a_t \in \mathcal{A}} \mathbb{E}[Q(s_t, a_t)]]$. It is attempting to find a policy $\pi_{s_{t+1}}(a_t, s_t)$ that produces the best costs and changes only by a certain amount from an action-selection prior $p(a_t)$.

For this term, the divergence bound between states and actions is assumed to be non-zero. States therefore carry some information about what action should be chosen, which allows for the formation of a non-uniform policy. In this instance, the magnitude of this second term $\inf_{\pi_{s_{t+1}}(a_t, s_t)} \mathbb{E}[\inf_{a_t \in \mathcal{A}} \mathbb{E}[Q(s_t, a_t)]]$ will be larger than the first $\inf_{a_t \in \mathcal{A}} \mathbb{E}[Q(s_t, a_t)]$. If the bound is zero, then $\inf_{\pi_{s_{t+1}}(a_t, s_t)} \mathbb{E}[\inf_{a_t \in \mathcal{A}} \mathbb{E}[Q(s_t, a_t)]] = \inf_{a_t \in \mathcal{A}} \mathbb{E}[Q(s_t, a_t)]$ and the value of information is zero.

As the number of training episodes becomes unbounded, the agent will have complete knowledge of the environment, assuming that the complexity-control parameter $\varphi_{\text{inf}}$ is equal to the state random variable entropy $\varphi_{\text{inf}} = -\sum_{s_t \in \mathcal{S}} p(s_t) \log(p(s_t))$. This second term will eventually produce returns that converge to those from the full-information case. The agent's behavior becomes entirely deterministic in such a situation, as the policy is a delta function for each state. Otherwise, the agent's behavior will be semi-random.

The constraint term in (3.2) can be described as follows:

**Constraint Term: Information Bound.** The policy complexity is determined by a Shannon information constraint $\mathbb{E}[D_{\mathrm{KL}}(\pi_{s_{t+1}}(a_t, s_t) \| p(a_t))]$. This constraint determines how much information the states carry about the actions. This term is bounded above by $\varphi_{\text{inf}}$, which implies that the mutual dependence between the states and actions will be artificially limited.

Naturally, the benefit yielded by the received information is related to the reduction in costs. As $\varphi_{\text{inf}}$ is increased, the states carry more information about what actions should be taken. The agent costs, as defined by the difference in the first and second terms, are expected to monotonically decrease. The converse is true for decreasing values of $\varphi_{\text{inf}}$. The costs are expected to increase, as there is less information about what action should be taken in a given state.

The constraint parameter $\varphi_{\text{inf}}$ has an impact on the amount of exploration. As $\varphi_{\text{inf}}$ decreases toward zero, the available search space is explored coarsely, and the agent begins to exploit more often than it explores. The size of the jumps that the policy iterates take in neighborhoods of the search space can increase compared to higher values of $\varphi_{\text{inf}}$. The frequency of the jumps decreases compared to higher values of $\varphi_{\text{inf}}$. When $\varphi_{\text{inf}}$ is zero, exploration becomes a random walk. The agent may therefore take an inordinate amount of time to complete an objective. When the parameter $\varphi_{\text{inf}}$ is set near to or above the state entropy, the action-state space is finely quantized. The agent therefore attempts to maximize the possible returns. The policy iterates move around frequently in small regions of the policy space.

The optimization of (3.1), under the constraint (3.2), can be performed by first converting the criterion into an unconstrained problem using the theory of Lagrange multipliers. Differentiating the unconstrained criterion and solving for the policy probabilities leads to an alternating, expectation-maximization-type update [35]. These alternating updates yield a soft-max-based action-selection process [1] for exploring the policy search space. For this style of exploration, the preference for one action over another is dictated by a Boltzmann distribution: actions with better returns are associated an increased preference chance. The final action choice is chosen at random using those preferences. It is well known that this type of weighted random exploration often empirically outperforms purely random strategies for many problems.

The action-selection process induced by using the value of information is not equivalent to soft-max exploration, though. The value of information is implicitly partitioning the state space according to the action-value function [36]. Partitioning occurs whenever the constraint parameter is below the state variable entropy. This quantization can make the search process more computationally efficient than other common exploration strategies. This is because similar states are grouped together and treated as a single state for the purposes of exploration. Fewer states therefore need to be investigated. It becomes possible to uncover high-performing policies early during learning, as a consequence. In contrast, no state grouping is performed for the conventional search heuristics in reinforcement learning.

### 3.2 Value of Information: Guided Exploration Case

Exploration should be deterministically driven to target regions of the policy space where the environment dynamics, as measured by the next-state transitions, are poorly understood. This helps ensure that high-performing policies, which might exist in those regions, can be found during training.

The value of information lacks the capability to recognize and investigate these types of regions, however. This is because actions are chosen solely in a weighted-random fashion during learning. Such a process does not account for the uncertainty in the transition dynamics, just the expected cost associated with each action. There may hence be large portions of the policy space that remain unexplored during the early learning process, unless the space is very coarsely quantized.

The value of information can be augmented so that it more effectively visits such regions. We do this with the inclusion of a model-based state transition component into the criterion. We would like for the agent to visit states and take actions that not only lead to the best reduction in average costs, but also yield a more complete model of next-state transitions. By attempting to maximize the transition model's predictive capability, the agent is compelled to consider all possible actions, not just those that lead to good costs. Seemingly sub-par actions, let alone actions that may not have been chosen at all, are taken if they improve the model's performance. This guides the state-action iterates into areas of the policy space where the transitions are currently poorly modeled, as was illustrated by Emigh et al. [9]; such areas often have not been visited before.

When dealing with completely observable, Markovian environments, a measure of the captured information about the agent's future is the Shannon mutual information carried by the action and the current state about the next state. Maximizing the mutual information ensures that the agent will choose actions that reduce the prediction uncertainty about future states in an actions-state sequence. This term can be incorporated into the value of information objective function as follows: we replace (3.1) by

$$\underbrace{\sup_{\pi_{s_{t+1}}(a_t,s_t)}\left(\sum_{s_{t+1}\in\mathcal{S}} p(s_{t+1})\log(p(s_{t+1})) - \sum_{s_t,s_{t+1}\in\mathcal{S}}\sum_{a_t\in\mathcal{A}} p(s_t)\pi_{s_{t+1}}(a_t,s_t)p(s_{t+1}|s_t,a_t)\log\left(\frac{p(s_{t+1},s_t,a_t)}{p(s_t)}\right)\right)}_{\sup_{\pi_{s_{t+1}}(a_t,s_t)}\left(I(s_{t+1};s_t,a_t)\right) = \sup_{\pi_{s_{t+1}}(a_t,s_t)}\left(H(s_{t+1})-H(s_{t+1}|s_t,a_t)\right)}. \tag{3.3}$$

We also replace (3.2) by information and cost constraints, respectively,

$$\pi_{s_{t+1}}(a_t,s_t) \text{ such that : } \begin{aligned} &\underbrace{\sum_{s_t\in\mathcal{S}} p(s_t)\sum_{a_t\in\mathcal{A}}\pi_{s_{t+1}}(a_t,s_t)\log\left(\frac{\pi_{s_{t+1}}(a_t,s_t)}{p(a_t)}\right)}_{\mathbb{E}\left[D_{\mathrm{KL}}(\pi_{s_{t+1}}(a_t,s_t)\|p(a_t))\right]} \leq \varphi_{\mathrm{inf}},\\ &\underbrace{\inf_{a_t\in\mathcal{A}}\left(\sum_{s_t\in\mathcal{S}} p(s_t)Q(s_t,a_t)\right) - \inf_{\pi_{s_{t+1}}(a_t,s_t)}\left(\sum_{s_t\in\mathcal{S}}\sum_{a_t\in\mathcal{A}} p(s_t)\pi_{s_{t+1}}(a_t,s_t)Q(s_t,\pi_{a_t'}(s_t))\right)}_{\inf_{a_t\in\mathcal{A}}\mathbb{E}\left[Q(s_t,a_t)\right] - \inf_{\pi_{s_{t+1}}(a_t,s_t)}\mathbb{E}\left[\inf_{a_t'\in\mathcal{A}}\mathbb{E}\left[Q(s_t,a_t')\middle|s_t\right]\right]} \leq \varphi_{\mathrm{cost}}. \end{aligned} \tag{3.4}$$

These constraints are parameterized by positive, user-selectable values $\varphi_{\text{inf}}$ and $\varphi_{\text{cost}}$. The value of $\varphi_{\text{inf}}$ artificially limits how much information the states should carry about what actions should be taken, while the value of $\varphi_{\text{cost}}$ artificially limits the obtainable cost for a given policy.

We refer to (3.3) and (3.4) as the uncertainty-based value of information. The interplay between each of the terms can be described as follows:

**First Term: Predictive Uncertainty.** The value of information was previously defined as the greatest reduction of costs associated with a complexity-constrained policy. The only term in the criterion now measures the predictive capability of the current action-state pair on the future state: $\sup_{\pi_{s_{t+1}}(s_t,a_t)} I(s_{t+1}; s_t, a_t) = \sup_{\pi_{s_{t+1}}(s_t,a_t)} \mathbb{E}[D_{\text{KL}}(p(s_t, s_{t+1}, a_t) \| p(s_{t+1})p(s_t, a_t))]$.

Due to the supremum term, the criterion seeks to find a transition model with the greatest predictive capability where the action-selection policy has a specified return bound and policy complexity bound. That is, the criterion attempts to maximize the reduction in the uncertainty about the next state in an action-state sequence, subject to the constraints. Choosing actions with the highest uncertainty has the effect of maximizing exploration, in the sense of either visiting the states or experiencing the state transitions, both of which effectively learn the state transition model.

The two constraints in (3.4) can be described as follows:

**First Constraint Term: Action Cost Bound.** The first constraint specifies the expected difference in returns for the no-information $\inf_{a_t \in \mathcal{A}} \mathbb{E}[Q(s_t, a_t)]$ and partial information $\inf_{\pi_{s_{t+1}}(a_t,s_t)} \mathbb{E}[\inf_{a_t \in \mathcal{A}} \mathbb{E}[Q(s_t, a_t)]]$ cases. These returns are bounded by $\varphi_{\text{cost}}$. As $\varphi_{\text{cost}}$ approaches negative infinity, we seek the simplest policy, as defined by $\varphi_{\text{inf}}$, with the lowest possible cost. For this to happen, the policy space must be searched finely. Exploration will therefore dominate. As $\varphi_{\text{cost}}$ approaches zero, the constraint to find the simplest policies, as defined by $\varphi_{\text{inf}}$, with the best returns is relaxed. The policy space is searched coarsely and exploitation dominates. For some value of $\varphi_{\text{cost}}$ between these two extremes, exploitation becomes more prevalent than exploration. The actual value for this to occur will be application-dependent.

**Second Constraint Term: Information Bound.** The second constraint $\mathbb{E}[D_{\text{KL}}(\pi_{s_{t+1}}(a_t, s_t) \| p(a_t))]$ measures the amount of information, in bits, obtained from the state random variable about the action random variable. The objective function therefore seeks to find a policy, with complexity below $\varphi_{\text{inf}}$, that has a corresponding return amount $\varphi_{\text{cost}}$.

The amount of policy simplification can have a profound impact on the resulting action-selection choices. If the compression amount is high, which corresponds to a policy of limited complexity, then the expected returns for a given policy may be low. This can imply that an agent may not be capable of executing enough actions to complete its objective. Alternatively, when the compression amount is low, the policy complexity will be negligible. The agent may therefore be capable of executing more actions than in the former situation. However, such a policy might not be parsimonious, and the agents may take unnecessary actions that can negatively impact the expected returns. For many problems, though, there is a clear region in which the trade-off between policy complexity and simplicity yields meaningful return improvements.

The optimization of (3.3), under the constraints in (3.4), can again be performed using the theory of Lagrange multipliers. This leads to an expectation-maximization approach for finding the action-selection probabilities. Since (3.3) is the optimization of Shannon mutual information term with respect to the conditional probabilities, the entire process is a convex optimization problem. Therefore, it is possible to find global minimizers of the criterion using this alternating approach.

### 3.2.1 Finding Policies with the Uncertainty-Based Value of Information

The expectation-maximization updates obtained by solving the uncertainty-based value of information criterion can be viewed as a soft-max style of exploration. This exploration process can be combined with Markov-decision-process-based $Q$-learning to perform reinforcement learning. This criterion can also be combined with other learning strategies, such as SARSA.

The combination of the value of information with $Q$-learning, for the discrete, tabular case, is given in algorithm 1. Although we are constructing a model for the transition dynamics, we do not utilize this model to implement model-based reinforcement learning. We instead opted for a model-free approach. Our reasoning is that a great many episodes may be needed for this transition model to become accurate enough to be useful for reinforcement learning. The issue is exacerbated if the states are described by high-dimensional features. This is because we rely on density estimation techniques to find the transition distributions, which suffer from the curse of dimensionality.

The corresponding optimization steps for the value of information are given in algorithm 1, steps 6 and 7. The update in step 7 leads to a process for weighting action choices in each state according to a Boltzmann distribution parameterized by the expected cost. This resembles soft-max-based action selection. The difference is that two extra

**Algorithm 1:** Value-of-Information-Based $Q$-Learning

1 Choose a non-negative values for the step-size $\alpha$, discount factor $\gamma$, agent risk-taking parameter $\theta$, and cost-influence parameter $\beta$.

2 Initialize the action-value function $Q(a,s)$, for all $s \in \mathcal{S}$, $a \in \mathcal{A}$.

3 **for** $t = 0, 1, 2, \ldots, T$ **do**

4 For all relevant $s_t \in \mathcal{S}$ and $a_t \in \mathcal{A}$, update the state probabilities
$$p(s_t) \leftarrow p(s_0) \sum_{s_0 a_0, \ldots, s_{t-1} a_{t-1}} p(s_1|a_0, s_0)\pi_{s_1}(a_0, s_0) \ldots$$

5 **for** $k = 0, 1, 2, \ldots$ until convergence **do**

6 For all relevant $s_t \in \mathcal{S}$ and $a_t \in \mathcal{A}$, update the action probabilities $p^{(k)}(a_t) \leftarrow \sum_{s_t \in \mathcal{S}} \pi^{(k)}_{s_{t+1}}(a_t, s_t) p(s_t)$.

7 For all relevant $s_t \in \mathcal{S}$ and $a_t \in \mathcal{A}$, update the action-selection probabilities $\pi^{(k+1)}_{s_{t+1}}(a_t, s_t) \leftarrow$

8 $$p^{(k)}(a_t)\, \omega_t \left( \textstyle\sum_{a_t \in \mathcal{A}} p^{(k)}(a_t)\, \omega_t \right)^{-1}, \text{ where } \omega_t = \exp\left( \left( \beta_t Q(s_t, a_t) + D_{\mathrm{KL}}(\pi^{(k)}_{s_{t+1}}(a_t, s_t) \| p(s_{t+1})) \right) \Big/ \theta_t \right).$$

9 Choose an action $\pi^{(k)}_{s_{t+1}}(a_t, s_t) \rightarrow a_t$ and perform a state transition $s_t \rightarrow s_{t+1} \in \mathcal{S}$. Obtain a cost $g_{t+1} \in \mathcal{R}$.

10 For $s_t \in \mathcal{S}$ and $a_t \in \mathcal{A}$, update $Q(a_t, s_t) \leftarrow Q(s_t, a_t) + \alpha_t \left( g_{t+1}(s_t, a_t) + \gamma_t \sup_{a \in \mathcal{A}} Q(s_{t+1}, a) - Q(s_t, a_t) \right)$.

terms have been included to account for the promotion or suppression explorative behaviors. These terms, along with associated parameters, weight the expected costs according to the level of exploration that an investigator wishes an agent to possess.

For step 7, there is an issue of how actually construct a transition probability model. We use kernel density estimators for this purpose. Such estimators need to be modified so that they are appropriate for transition probabilities. We follow the ideas outlined by Emigh et al. [9] for this purpose. We assume that neighboring states have similar transition probability distributions and weight the contribution of each observation by a similarity measure. We also assume that transition distributions are centered at the point of transition. Analogous constraints are chosen for the actions. We then specify a density estimator for the transitions that relies on a product of kernels for handling state similarity, action similarity, and rewards to implement these assumptions [9]. The divergence between successive models can be computed using information-theoretic concepts [6], which provide closed-formed solutions that are efficient to compute.

The parameters $\beta_t$ and $\theta_t$ that arise upon minimization of the unconstrained value of information dictates the rate of change in (3.3) with respect to the constraints in (3.4). Their effects can be describes as follows:

**Cost-Influence Hyperparameter.** The parameter $\beta_t$ dictates the cost emphasis on the action-selection process. As $\beta_t$ goes to zero, the agent increasingly does not prioritize actions that reduce costs. Actions are instead chosen that solely attempt to maximize the predictive power of the transition-probability model. This behavior can be useful early in the learning process, as regions of high-performing policies can be quickly uncovered. The emphasis on prediction-model uncertainty decreases for increasing $\beta_t$. As $\beta_t$ becomes infinite, actions are chosen that completely reduce costs. This behavior can be useful later in the learning process, as we would like for the agent to refine its task-completion strategies once the policy space has been semi-uniformly investigated.

**Information-Influence Hyperparameter.** The parameter $\theta_t$ controls the determinism of the action-selection process. As $\theta_t$ goes to zero, the selected action will attempt to simultaneously maximize the transition-model divergence and the action-value function maximization. In this situation, the policy complexity constraints are ignored and a fine-grained exploration of the policy space is performed. This implies that the action-selection process will be highly random. As $\theta_t$ becomes infinite, the relative entropy between the actions and states is minimized. This implies that the probability of choosing an action based upon the current state is independent of the current state. That is, all states elicit the same response, since the policy is so simple that the agent cannot change its behavior. The action-selection process is therefore deterministic. Despite this limitation, the agent will greedily select actions that attempt to optimize the expected policy returns.

For values of $\theta_t$ between these two extremes, there is a trade-off that occurs. As the parameter value increases, the agent may favor state-specific actions that yield a better expected return for particular states. This leads to a high relative entropy between the states and actions, implying that the policy complexity will be high. Alternatively, as the parameter value decreases, the agent may favor actions that yield a good expected utility for many observations, which produces a lower relative entropy. The resulting policy complexity will hence be lower. Regardless of the value $\theta_t$, exploration is performed. This contrasts with the original version of the value of information, where exploration would only be performed for non-zero values of this parameter. In such a case, the chosen action simply attempts to maximize the return; the selection is completely greedy.

It should be noted that the granularity of the search process changes only at critical values of $\theta_t$. For values of $\theta_t$ between two critical values, the number of state-space partitions remains constant. Critical values can be determined at those instances where the determinant of the Hessian is zero [37].

**Automated Hyperparameter Selection.** Choosing good parameter values for $\beta_t$, and $\theta_t$ can be challenging, as their influence over the exploration process is application dependent. There are a few strategies that can be used for this purpose, however. One example would involve using expert knowledge. Investigators often have expectations about the scores they wish for the policies to obtain. A reasonable upper and lower bound for these scores could then be used to derive corresponding parameter values. The parameter values would then be fixed during the learning process, which leads to an unchanging amount of risk. Another example would be to tune the parameter values while the agent is learning. This could be done, for instance, by annealing. Assuming a sufficiently slow rate of change, properties of simulated annealing indicate that the parameters will eventually settle at values that yield near-optimal-cost policies.

An alternate procedure that we have found to work well involves the use of the policy cross-entropy. When applied to reinforcement learning, cross-entropy

$$\mathbb{E}_{\pi^{(k)}_{s_{t+1}}(a_t,s_t)}\left[-\log\left(\pi^{(k+1)}_{s_{t+1}}(a_t,s_t)\right)\right] = -\sum_{s_t\in\mathcal{S}} \pi^{(k)}_{s_{t+1}}(a_t,s_t)\log\left(\pi^{(k+1)}_{s_{t+1}}(a_t,s_t)\right) \tag{3.5}$$

quantifies the amount of overlap between two policies. It is a bounded measure of the amount of change between policies when acting on policies $\pi^{(k)}_{s_{t+1}}(a_t,s_t)$ and $\pi^{(k+1)}_{s_{t+1}}(a_t,s_t)$ across consecutive episodes. This reflects the rate of learning about the environment dynamics. If the action-selection probabilities change by a great amount, then the cross-entropy will be near one. The acquisition of new behaviors can therefore be prevalent, which typically precedes a steep decrease in cost. If action-selection probabilities change only slightly, then the cross-entropy will be near zero. Existing behaviors are likely to be retained, leading to moderate or negligible improvements in costs.

Policy cross-entropy can be employed to yield an efficient hyperparameter adjustment scheme when using uncertainty-based value-of-information exploration. A low action-state complexity should be used, in the beginning, to sparsely survey the entire solution space. High-performing regions will be quickly uncovered when doing so, as the state-action space will be coarsely quantized. The policy tables will begin to be populated with meaningful entries, and basic gameplay behaviors will emerge. Eventually, the cross-entropy will dwindle and reach a near-steady-state value. Before this occurs, the policy complexity should be gradually increased to promote a more thorough investigation of the policy space. The action-state space quantization will be relaxed, allowing existing action strategies to be refined. Further cost improvements can often be realized.

## 4 Centipede Simulations

We first assess our uncertainty-based exploration approach on the classic arcade game Centipede. Learning good policies for this game is challenging: there are a variety of ways the agent can score points, some of which are better than others but carry increased risks.

The aims of our simulations with this game are two-fold. First, we want to demonstrate the influence of both hyperparmeters on the search process. This will be done through an analysis of both the state-visitation trajectories and the action-value function estimates. We consider a simplified state space, so as to visualize and analyze the learning process. Results are provided in section 4.2.1. Secondly, we want to quantify the performance of our approach, as dictated by the obtainable cost. We compare these costs to those from both uncertainty-based and stochastic exploration schemes. These comparisons are given in section 4.2.2.

### 4.1 Simulation Preliminaries

#### 4.1.1 Agent Gameplay and Rewards

In the game Centipede, the player dictates the lateral movement of an agent. The objective of every stage is to eliminate all of the centipede's body segments by having the agent shoot laser bolts at them. Destroying part of the centipede's body results in a small negative cost ($-10$), while destroying the head results in a larger cost ($-100$).

This simple objective is impeded in several ways. First, there are several mushrooms present in the environment, which act as barriers. Mushrooms can absorb several laser bolts before disappearing and yield a minuscule negative cost when this occurs ($-1$). Mushrooms can make the centipede segments hard to hit with laser bolts from certain positions. As well, shooting any section of the centipede creates a new mushroom. Randomly spawned enemies known as fleas have the ability to leave additional mushrooms in their path.

Mushrooms have other effects on the gameplay dynamics. When the centipede comes in contact with one, it drops one row closer to the agent and switches direction. Thus, more mushrooms in the environment cause the centipede to descend more rapidly, placing the agent in danger of losing a life. Mushrooms can also be poisoned by randomly spawned scorpion enemies. If the centipede encounters a poison mushroom, then it hurtles straight down toward the player. The centipede returns to its normal behavior upon reaching the bottom row of the environment.

The agent loses a life when it is hit by either a centipede or another enemy, such as a spider or a flea. When a life is lost, any partially damaged mushrooms are completely regenerated; poisoned mushrooms revert to normal mushrooms. Points are awarded for each regenerated mushroom ($-5$). Likewise, destroying fleas ($-200$), spiders ($-300$ to $-900$), or scorpions ($-1000$) leads to large negative costs. A game ends if all of the agent's lives are gone. A good policy should therefore attempt to minimize the total cost without losing lives too quickly.

#### 4.1.2 Extracted Features

There are a variety of features that could be employed to produce high-performing agents. We are, however, interested in visualizing the exploration process, which artificially constrains the number of features. We hence consider just two holistic features: the current lateral position of the agent and the lateral position of either the closest enemy or the closest centipede segment to the agent. Both of these features, surprisingly, furnish the agent with a great deal of potential insight into the gameplay dynamics.

### 4.2 Simulation Results and Discussions

Simulations of the Centipede game take place in a discrete, two-dimensional environment composed of $30 \times 30$ grid cells. In the original Centipede game, the number of horizontal grid cells was increased, which allowed for more precise agent movement. We artificially reduced the number of horizontal grid cells so that other uncertainty-based exploration approaches could be applied.

We have four free parameters that need to be set during the learning process. Two of these, the cost-influence parameter $\beta_t$ and the information-influence parameter $\theta_t$, are adjusted throughout our simulations. The remaining two parameters are the learning rate and the discount factor, which dictate the action-value update behavior when using $Q$-learning. We set the learning rate $\alpha$ to 0.45 so that more weight is given to previously acquired action-value magnitudes than those that were more recently acquired. We temporarily increase the learning rate to 0.7 for a single episode if that episode has the lowest cost compared to all previous episodes. This ensures that the policy quickly implements new agent behaviors to better play the game. A discount factor $\gamma$ of 0.9 was used so that the agent would seek action sequences with low long-term costs. Each of these values was chosen based upon our previous applications of $Q$-learning to stochastic game environments.

#### 4.2.1 Hyperparameter Effects

In this subsection, we analyze the effects of the cost-influence $\beta_t$ and information-influence $\theta_t$ hyperparameter values on the exploration process and hence the obtainable cost. We also comment on the changes in agent behaviors.

The first hyperparameter that we assess, $\beta_t$, specifies the importance of costs when weighting the action choices. It hence impacts exploration. We consider five fixed values of $\beta_t$ for all episodes $t$: 0.1, 0.25, 0.5, 1.0, and 2.0. We also performed additional simulations with values between these five examples to augment the discussions. We set the information-influence hyperparameter, $\theta_t^{-1}$, to 0.25. This value of $\theta_t^{-1}$ corresponds to a somewhat moderate quantization of the action-state space.

The information-influence hyperparameter, $\theta_t$, also dictates how finely the state-action space will be quantized and hence how thoroughly it will be investigated. We consider five fixed values of $\theta_t^{-1}$ for all episodes $t$: 0.1, 0.25, 0.5, 0.75, and 1.0. We fix the cost-influence hyperparameter, $\beta_t$, at 1.0 for these different values of $\theta_t$. This value of $\beta_t$ indicates that costs are equally important as the transition model quality when weighting the possible actions.

**Policy Cost and Learning Rate Results.** Results for the simulations in this section are given in figure 4.1. Plots of the average accrued agent costs for the five hyperparameter choices of $\beta$ and $\theta_t$ are provided in figures 4.1(e) and 4.1(f), respectively. Lower costs imply better agent performance. We also include plots of the average policy learning rates in figures 4.1(e) and 4.1(f) for the case where $\beta_t$ and $\theta_t$ change, respectively. The policy learning rate was was derived from the cross entropy of the actions-selection probabilities across consecutive episodes. Lower learning rates imply that fewer changes are being made to the policy; it is less likely that the policies can account for novel environment dynamics when this occurs. These averages were obtained over 100 Monte Carlo trials.

As shown by the left plot in figure 4.1(e), the choice of $\beta_t$ has a profound impact on the learning process. Small values of $\beta_t$, such as $\beta_t$ equal to or below 0.1, led to poor long-term costs. This is despite that the policy learning rates were somewhat high, as highlighted in right plot of figure 4.1(e). These high learning rates signified that the policy was changing rather significantly across episodes to better account for the environment dynamics; low costs should be obtained, but this did not occur. As $\beta_t$ was increased, the obtainable costs improved, up to a certain value for $\beta_t$ between 1.0 and 2.0. Past this threshold, the agent costs did not improve compared to when $\beta_t$ was approximately 1.0. The corresponding policy learning rates were also low after so many episodes, which indicated that the policies were not changing greatly. Increasing $\beta_t$ beyond 2.0 had only a negligible impact for the chosen value of $\theta_t$. Either increasing $\theta_t$ from 0.25 to 0.75 or decreasing $\theta_t$ from 0.25 to 0.05, in increments of 0.01, also did little to change the outcomes.

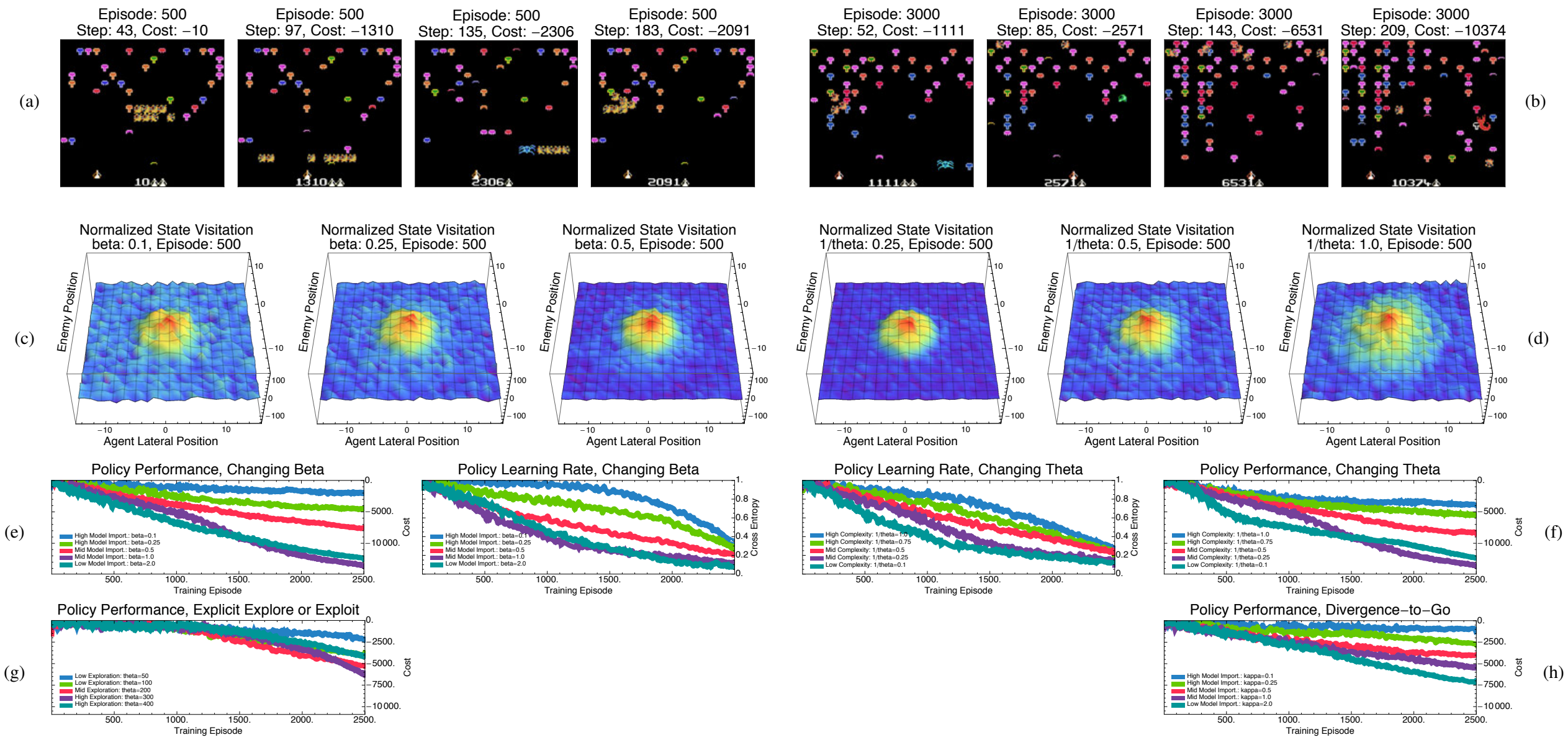

Figure 4.1: Results of the agent's performance during training for the game Centipede. In rows (a) and (b) are snapshots of the agent's gameplay behaviors. Early during training, as shown in (a), the agent performs many unnecessary actions in an attempt to understand the environment dynamics. It will rarely attack enemies and usually loses a life due to the centipede reaching the bottom of the screen. Later in training, the transitions are better understood. As indicated by the costs in (b), the agent can consistently move to attack the nearest enemies by this point of the learning process.

Quantitative results of the learning process are provided in (c)–(h). In rows (c) and (d) are surface plots of the averaged state visitations when changing the cost-influence hyperparameter $\beta_t$ and the information-influence hyperparameter $\theta_t$, respectively. For these plots, the height of the surface indicates which states are visited frequently. These plots show that the center of the state space is visited often, regardless of the parameter values, as that is the only state where the agent can fire its laser at enemies and reliably score points. In rows (e) and (f) we plot the policy costs and learning rate when changing $\beta_t$ and $\theta_t$, respectively, for the uncertainty-based value of information. Each colored curve corresponds to a unique value of either $\beta_t$ and $\theta_t$. Lower costs indicate a better-performing policy. Lower cross-entropy values indicate fewer changes being made to the policy. Better costs are obtained for high values of $\beta_t$ and low values of $\theta$. Lastly, in rows (g) and (h) are the policy costs produced when using the explicit-explore-or-exploit algorithm and divergence-to-go, respectively. Neither of these methods approaches the performance of the uncertainty-based value of information, regardless of the chosen parameters.

The plots in figure 4.1(f) capture the effects of the information-influence hyperparameter, $\theta_t$, on the performance. For small values of $\theta_t$, the agent costs tended to drop quickly during the first few hundred episodes. The costs continued to improve over the next two thousand episodes. After around three thousand episodes, the costs reached steady-state values. This decrease in the cost change coincided with the decline in the policy learning rates shown in the left-hand plot of figure 4.1(f). Such values yield a coarse quantization of the environment. Increases in $\theta_t$, past around 0.25, led to worse costs. Values of $\theta_t$ higher than 0.25 correspond to rather finely grained quantizations of this environment, which do not appear to be necessary for good performance in this domain. That is, a high amount of exploration did not seem to be necessary for this simple state-action space.

**State Visitation Results.** To better examine the influence of $\beta_t$ and $\theta_t$ on the exploration process, we provide, in figures 4.1(c) and 4.1(d), graphs of the normalized state-visitation densities at various phases of the learning process. For each of these plots, states that are visited more often are assigned a higher magnitude and denoted using a warmer color. States in red correspond to frequently visited states. In contrast, states that are visited infrequently have a low magnitude and are represented using a cooler color. States shaded in purple are not visited much, if at all.

The plots in figure 4.1(c) illustrate that, when $\theta_t$ was fixed, low values of $\beta_t$ led to a thorough exploration of the state space. Actions were chosen such that many states in the environment are visited, albeit at a detriment to costs. That is, the action-value function magnitudes were rather low, which aligns with the costs displayed in figure figure 4.1(e). When $\beta_t$ was raised, the amount of times certain states were visited was reduced. Certain regions of the state space were also not visited, unlike when lower values of $\beta_t$ were used. The action-value magnitudes did improve, however, for higher values of $\beta_t$. Again, past a certain threshold, higher values of $\beta_t$ did not lead to meaningful improvements of the action-value magnitudes for the value of $\theta_t$ we utilized.

Changing $\theta_t$ also has the effect of adjusting the amount of exploration. For $\theta_t$ approaching zero, the policy space was coarsely quantized and sparsely searched. This is captured by the plots in figure 4.1(d). Good action-value function magnitudes were obtained in this case. As $\theta_t$ approached its maximal value, the agent increasingly selected actions at random, regardless of their expected long-term costs. Consistently taking random actions had the effect of investigating the state space thoroughly, albeit unnecessarily for this simple problem. This is shown in the right-most plot in figure 4.1(d). A similar outcome was witnessed when $\beta_t$ was decreased toward zero. The difference, however, is that the exploration was deterministically driven by $\beta_t$. Actions were consistently assigned a high probability of being chosen so that the predictive model quality would improve.

**Hyperparameter Effects Discussions.** Through the preceding simulations, we demonstrated the effects of both $\beta_t$ and $\theta_t$ on the agent's performance. Both hyperparameters influence the exploration process.

Our results indicate that low, fixed values of $\beta_t$ led to poor long-term costs for the chosen value of $\theta_t$. This was expected. As we indicated indicated in the previous section, $\beta_t$ scales the action-value magnitude for each state-action pair. Small values of $\beta_t$ hence de-emphasize the contribution of the action-value function, which implies that expected costs become mostly irrelevant when weighting the action choices. Only the current- and next-state divergence, and hence the model's prediction quality, matter. The exploration process was therefore focused solely on building a better state-transition model. We observed this behavior in the state-trajectory plots: states were visited several times in an attempt to experience all possible transitions.

These values of $\beta_t$ had a profound influence on the agent's behaviors. Often, the agent would shy away from attacking enemies in favor of moving to an adjacent grid cell. This occurred even when the agent could readily attack enemies that would greatly decrease the total cost, such as spiders, fleas or scorpions, as shown in figure 4.1(a). The agent would also fire randomly, even when no enemies were close enough for it to attack. Such behaviors were, as we noted above, attempts to understand the influence of the chosen actions on the next-state transitions. However, these types of gameplay strategies were largely unnecessary for understanding the environment and reducing long-term costs: the environment size was small and the transition dynamics were rather simple.

A great many episodes were wasted for these low values of $\beta_t$, as indicated by the stalled costs early in the learning process. This was despite the rather high policy learning rates, according to the policy cross-entropy, throughout training. We found, however, that high learning rates were a byproduct of taking many sub-optimal-cost actions and hence state transitions. That is, after experiencing such transitions, many of the probabilistic policy entries would consistently change by small amounts to reflect the corresponding action's immediate influence on the transition-model uncertainty. The aggregation of these small changes led to a moderately high cross-entropy. Hence, while the policy was being changed to better capture the environment dynamics, doing so was not actually beneficial for good long-term performance in this simple problem domain.

Higher values of $\beta_t$ fared better for this game, up to a point. Such values signaled that the absolute magnitude of the action-value function was important for deciding which actions should be taken. That is, actions that lead to low, long-term expected costs should be chosen over those that do not; the influence of the transition-model uncertainty was thus lessened. This change in $\beta_t$ ensured that the agent would move to attack enemies whenever feasible, as shown in figure 4.1(b). It would also either dodge or fire at those enemies that could easily harm it, such as spiders. Both behavior changes helped the agent to accrue a lower cost. It also allowed the agent to survive much longer. While some exploration was necessary to develop such cost-reducing behaviors, this could be effectively handled via the stochastic exploration mechanism, as dictated by $\theta_t$.

For $\beta_t$ above one, we did not witness any long-term improvement in the obtainable costs compared to other values of $\beta_t$. This was expected. When increasing $\beta_t$ too much, the effects of the action's costs outweigh the potential improvement in the prediction model uncertainty. Actions are more likely to be selected stochastically, which leads to a random walk in the state-action space. The exploration process therefore will not deterministically investigate unvisited regions of the joint space. Even for simple problem domains, like this one, a guided exploration can help refine the action-selection strategy.

The hyperparameter $\theta_t$ governs the randomness of the exploration process. As we showed in [7], it also regulates the risk in the utility change that the agent is willing to tolerate when choosing an action. For this problem domain, low amounts of risk were beneficial for good long-term performance. That is, once an agent had uncovered a reasonably good policy, it could follow that policy without modifying it much. Higher amounts of risk implied that seemingly cost-sub-optimal actions should be chosen at random and consistently taken during training. With so few states and actions in this problem, though, deviating from a reasonably good action was often not beneficial. More complicated problem domains would benefit from higher values of $\theta_t$, however, which we show in the next section.

This hyperparameter also implicitly yields a fuzzy quantization of the underlying Markov chain transition matrix according to the action-value function, which follows from our work in [36]. It hence is partitioning the state space and assigning a common set of action-selection probabilities to each state group. For this problem domain there are many states that can be grouped together. For example, if the agent is either to the right or to the left of the nearest enemy, then it should either move to the left or to the right, respectively. If the agent is below the nearest enemy, then it should shoot. Very little action-specialization is hence necessary, and only a coarse quantization is needed. This further explains why low values of $\theta_t$ led to good results in these simulations: they induced a small number of state groups with a large number of states per group.

Given the importance of $\beta_t$ and $\theta_t$ on the exploration process, it is crucial that appropriate values be chosen. In the next section, we demonstrate that these hyperparameters can be automatically tuned according to our cross-entropy heuristic. This avoids the need to run batteries of simulations for each problem domain in an attempt to manually arrive at good values.

#### 4.2.2 Methodological Comparisons

In this subsection, we provide context for the above results. We compare the results of our approach with two other uncertainty-based exploration schemes: the explicit-explore-or-exploit algorithm of Kearns et al. [12, 13] and the divergence-to-go method of Emigh et al. [9]. Both approaches address the exploration-exploitation dilemma in different ways.

**Explicit-Explore-or-Exploit Algorithm Results.** The explicit-explore-or-exploit algorithm is one of the most well known uncertainty-based exploration schemes, which is due to its theoretical guarantees. The basic idea of this algorithm is that it attempts to approximate an environment's Markov decision process by another Markov decision process. That is, it repeatedly explores state-action pairs whose transition dynamics are still inaccurately modeled. After a certain number of state visitations, the approximate model is deemed to be accurate enough for those states, even if it is not true. Exploitation is then performed on such states to minimize the sum of costs over time.

We apply this algorithm to $Q$-learning and the same learning rate and discount factors as the uncertainty-based value of information. There is only one remaining hyperparameter value, $\vartheta_t$, that must be set. The value of $\vartheta_t$ dictates when a state has been visited a sufficient number of times and hence when exploration is no longer necessary. We consider five fixed values of $\vartheta_t$ for all episodes $t$: 50, 100, 200, 300, and 400 state visitations. Simulations with values between these five examples were also performed to enhance the discussions. Lower values of $\vartheta_t$ favor exploitation, while higher values lead to more exploration.

Plots of the average costs across 100 Monte Carlo trials are given in figure 4.1(g). As shown in this plot, the costs are worse, regardless of the hyperparameter value, than those for the uncertainty-based value of information. This was despite the high policy learning rates throughout the learning process. By conducting additional simulations, we found that deviating from the chosen hyperparameter values did little to improve the results. Only increasing the number of episodes had any meaningful effect, as it ensured each state would be visited enough times during learning to switch from exploration to exploitation. Double the number of episodes was needed compared to the uncertainty-based value of information to obtain the same costs, though.

**Divergence-to-Go Results.** The other uncertainty-based approach that we consider, divergence-to-go, is a modification of kernel temporal-difference learning. In particular, the algorithm replaces the action-value function with a next-state-transition model divergence. This has the effect of favoring actions that most quickly improve the transition model. The search process is therefore driven into regions of the environment where the dynamics are distinct from the dynamics surrounding areas, which is similar to the behavior of the uncertainty-based value of information.

Divergence-to-go has three free hyperparameters. The first two are the learning rate and discount factor, which are analogous to the same terms found in $Q$-learning. We therefore utilize the same values as for the uncertainty-based value of information. The remaining parameter $\kappa_t$ specifies the kernel bandwidth for the transition-model density estimation. We consider six fixed values of $\kappa_t$ across episodes $t$: 0.1, 0.25, 0.5, 1.0, 2.0. Simulations with values between these five examples were also performed to enhance the discussions. Increasing values of $\kappa_t$ lead to worse-performing transition models and a sparser exploration of the space.

We provide plots of the average agent performance in figure 4.1(h). Values of $\kappa_t$ below 0.25 led to poor costs and conducted a high amount of exploration over a finite horizon dictated by the discount factor $\gamma_t$. That is, the agent took actions that attempted to build a good transition model within a finite number of action choices. The policy learning rate was consistently high early during training, then dropped after a few hundred episodes, as indicated by the policy learning in figure 4.x(x). Conversely, values of $\kappa_t$ past 0.5 searched less thoroughly. This had the effect of improving costs compared to when $\kappa_t$ was small, as indicated by figure 4.1(h). Values of the kernel bandwidth $\kappa_t$ beyond 2.0 tended to explore much too coarsely to adopt cost-reducing behaviors for this game.

**Methodological Discussions.** The preceding simulations indicate that the uncertainty-based value of information outperforms both alternate approaches. In the case of the explicit-explore-or-exploit algorithm, this was to be expected. This approach requires that exploration be performed until an accurate model of the entire Markov decision process is obtained. Alternatively, it requires exploration for the reachable parts of the Markov decision process. The strong search bias of this approach requires revisiting many states that might already have accurate value-function approximations and hence provide a good indication of the resulting actions that should be taken. A great many learning episodes can be wasted, as a consequence.

To elaborate, this approach routinely investigated regions of the state-space that were associated with cost-poor actions. It would, for instance, elect to have the agent continuously move away from the nearest enemies. Rarely would the agent attack, in such cases. The agent would therefore often not improve its score throughout many learning episodes. This exploration approach would also compel the agent to remain stationary, which would leave it vulnerable to enemies that could reach the bottom of the level, such as spiders or fleas. It would also intentionally move so as to collide with these enemies, along with centipedes. Both action responses consistently cut short each learning episode compared to when the uncertainty-based value of information was employed. While such behaviors could prove helpful, in the long term, for understanding the gameplay dynamics, they were not helpful in the short term.

Such issues stem from the explicit-explore-or-exploit algorithm's sensitivity to the its single hyperparameter $\vartheta_t$. It is difficult to set $\vartheta_t$, since it is application dependent. That is, too low a value can lead to a shallow and ineffective survey of the environment, while too high a value can redundantly explore the environment. While the uncertainty-based value of information also has application-dependent hyperparameter values, our experiments on a wide range of game applications appear to indicate that the sensitivity is typically much less pronounced. Poor values for $\beta_t$ can often be counteracted by the influence of $\theta_t$ on the search process and vice versa. We also provide an automated means to set these hyperparameter values, unlike the explicit-explore-or-exploit algorithm or any of its extensions.

These issues also stem from explicit-explore-or-exploit algorithm's well-documented flaw that it cannot generalize well to classes of stochastic games [16, 17]. In many games, the agent does not directly control its adversary's behavior, it may only implicitly influence its response. Likewise, it may not be possible to accurately predict the adversary's behavior. It is therefore difficult to determine the outcomes of the agent's actions and whether or not they lead to new information. For the explicit-explore-or-exploit algorithm to practically perform well for such problems, it would need to characterize the information content of states and actions, which it does do. In contrast, the uncertainty-based value of information accounts for the informativeness of the actions, in the form of the state-transition model divergence. If a state transition was unexpected, then our approach will revisit this state until the model divergence is minimized and hence no more useful information can be obtained.

Divergence-to-go has a search mechanism resembling that of the uncertainty-based value of information. It assess the divergence of the predicted next-state transition and the actual transition that occurred to determine which action should be chosen the next time the agent encounters a particular state. Despite this similarity, divergence-to-go often underperformed compared to our approach, depending on the choice of $\kappa_t$. The reason was that divergence-to-go did not account for the expected agent costs when choosing actions. It merely attempted to explore so as to uncover the best state-transition model. There is no guarantee that constructing a good transition model will yield good costs for this particular application.

The agent behaviors implemented by divergence-to-go sometimes mirrored those of the explicit-explore-or-exploit algorithm. For low values of $\kappa_t$ between 0.01 and 0.5, the search process visited many of the states, even those that did not yield good costs. This was due to the low kernel bandwidth, which did not permit an effective generalization of the divergence information to nearby states. That is, many states needed to be visited several times to reliably estimate the divergence, which was due to the small generalization radius. Since the search process did not adequately visit gameplay-informing states for the limited number of episodes that we considered, the agent's behavior was erratic. It would often needlessly move laterally, along with dodging enemies that it could easily attack. The agent would also not take easy shots that would improve its accrued cost. For higher values of $\kappa_t$, the volume of the state space that was explored decreased. This, paradoxically, led to improved behaviors and costs compared to the explicit-explore-or-exploit algorithm. The agent would often track the nearest centipede segments and continuously shoot at it, even through mushrooms, before seeking out other nearby enemies to attack. It would also avoid spiders and fleas that came too close to the agent, which could cause a loss of life. Such behaviors emerged since the kernel bandwidth permitted the generalization of divergences from nearby states. Those states that were visited contained informative details about what happens after the agent moves toward an enemy and attacks.

## 5 Crossy Road Simulations

We now assess our exploration strategy on the more complicated game: Crossy Road. Learning good policies for Crossy Road is a highly challenging problem. Foremost, the game takes place in a stochastic, dynamic environment. As well, the game requires a shift in play styles throughout certain sections. Without an ability to adequately explore the state space, the development of good agent behaviors that can deal with various gameplay issues may not occur early in the learning process.

The aims of our simulations with this game are two-fold. First, we want to discern what agent behaviors emerge and understand the impact of the search processes on those behaviors. Results are presented in sections 5.2.1 and 5.2.2. Secondly, we want to quantify the performance of policies obtained with our uncertainty- and stochastic-based exploration strategy with those that are purely stochastic. This is done in section 5.2.3. For our purposes, performance will be measured according to cost and cross entropy.

### 5.1 Simulation Preliminaries

#### 5.1.1 Agent Gameplay and Rewards

Crossy Road is an endless dynamic maze game. The objective is for the agent to move forward through a semi-confined environment while avoiding collisions with any of the dynamic obstacles. Dynamic obstacles, such as automobiles and trains, appear on roadway and track tiles. Colliding with a dynamic obstacle results in a loss of a life. A loss of life can also occur if the agent either waits too long to move forward, moves backwards more than three rows, or jumps into a water tile. When the agent runs out of lives, the position of the agent is reset, and a new procedurally generated layout is created.

The scoring system for our implementation of Crossy Road is as follows. If a move leads the agent to a new row in the environment, a negative cost is incurred ($-100$). Taking any movement results in a positive cost ($+10$), as does waiting ($+10$), which helps insure that the agent does not remain unnecessarily stationary. A negative cost is added when the agent picks up any randomly spawning gold coins ($-500$). Red coins are worth even more ($-750$). These coins have no effect on gameplay; they simply unlock various agent customization options. Here, coins serve as secondary objectives to make the reinforcement learning problem more interesting. Colliding with a dynamic obstacle results in a high penalty ($+500$). Losing via other means is also penalized ($+300$). A good policy should attempt to minimize the total cost.

#### 5.1.2 Extracted Features

We consider two separate state-space feature sets for Crossy Road. This is done to ensure that our exploration mechanism works well in different contexts.

The first feature set is a hybrid mixture of local and holistic features. The local features provide spatial information about the grid occupancy three grid locations away from the agent in the forward direction. Grid occupancy is only provided two grid locations away for to the left, right, and rear of the agent. Information about diagonal grid occupancy is also provided. Such features outline which grid locations are safely reachable, contain a coins, or contain either static or dynamic obstacles. The holistic features describe more global information. One example is the direction of a traveling tree log, provided that the agent is currently on a log. Other examples are the existence and travel direction of any dynamic obstacles up to two rows away and the distances to the two nearest dynamic obstacles. In total, twenty-one features are used.

Using holistic with local features ensures that the agent does not aimlessly wander through the event. When using purely local features, such undesirable behaviors could materialize, as the agent can only make observations a few grid locations away. Increasing the observation radius for the local features would avoid such behaviors. It would also greatly increase the state-space size, though, and hence complicate training.

The second state-space feature set relies predominantly on holistic features. Such features include the quantized distance to the three nearest dynamic obstacles, the direction of those obstacles relative to the agent, and the direction that the obstacles are moving. We include attributes for assessing if the current and next two rows contain dynamic obstacles and the direction that those obstacles are traveling. The quantized distance and direction to the nearest coin is also used. We additionally provide information about the agent's direction of travel if it is on a moving log. Lastly, we have local features that provide information about the type and occupancy of the cells up to one grid location away from the agent. In total, twenty-one features are extracted. Like the hybrid features outlined above, the holistic features are insensitive to the spatial location of the agent in the environment and therefore should generalize across different levels.

### 5.2 Simulation Results

For our simulations, we consider a two-dimensional, grid-based level configuration composed of $15 \times \infty$ grid cells. These cells can be populated by static and dynamic obstacles. At any time, only a $9 \times 19$ subsection of this grid is visible. We impose the constraint that the obstacles cannot be positioned on a row such that the agent does not have a path forward.

Four free parameters must again be set. We consider different values for the information-influence parameter $\theta_t$ throughout our simulations, while keeping the cost-influence parameter $\beta_t$ fixed. For the learning rate $\alpha$ and the discount factor $\gamma$, we consider the same values as in the previous section.

#### 5.2.1 Fixed Hyperparameter Effects

In this subsection, we analyze the performance for this game environment when using fixed values of $\theta_t$ and $\beta_t$. This is done to establish a cost baseline for the cross-entropy-tuned results presented in the next section.

For $\theta_t$, we consider three different values: 0.1, 0.35, and 0.7. These values yield low-complexity, moderate-complexity, and high-complexity policies, respectively. For $\beta_t$, we utilize a value of 1.0. This value offers a reasonable compromise between deterministic exploration and the transition-model quality.

Our simulation results are provided by the plots in figure 5.1. Due to the high dimensionality of the feature space, we are unable to plot the state visitation trajectories. We therefore consider only average costs and the policy learning rates as described by cross entropy.

Figures 5.1(e) and 5.1(f) capture both the holistic and hybrid policy performance and learning rate across 10000 episodes for 100 Monte Carlo simulations. The results indicate an initial sharp decrease in cost for three policy complexities. This is due to an effective exploration of the search space and the exploitation of high-performing actions. Later in learning process, there is a tempering of the cost drop-off, depending on the employed policy complexity.

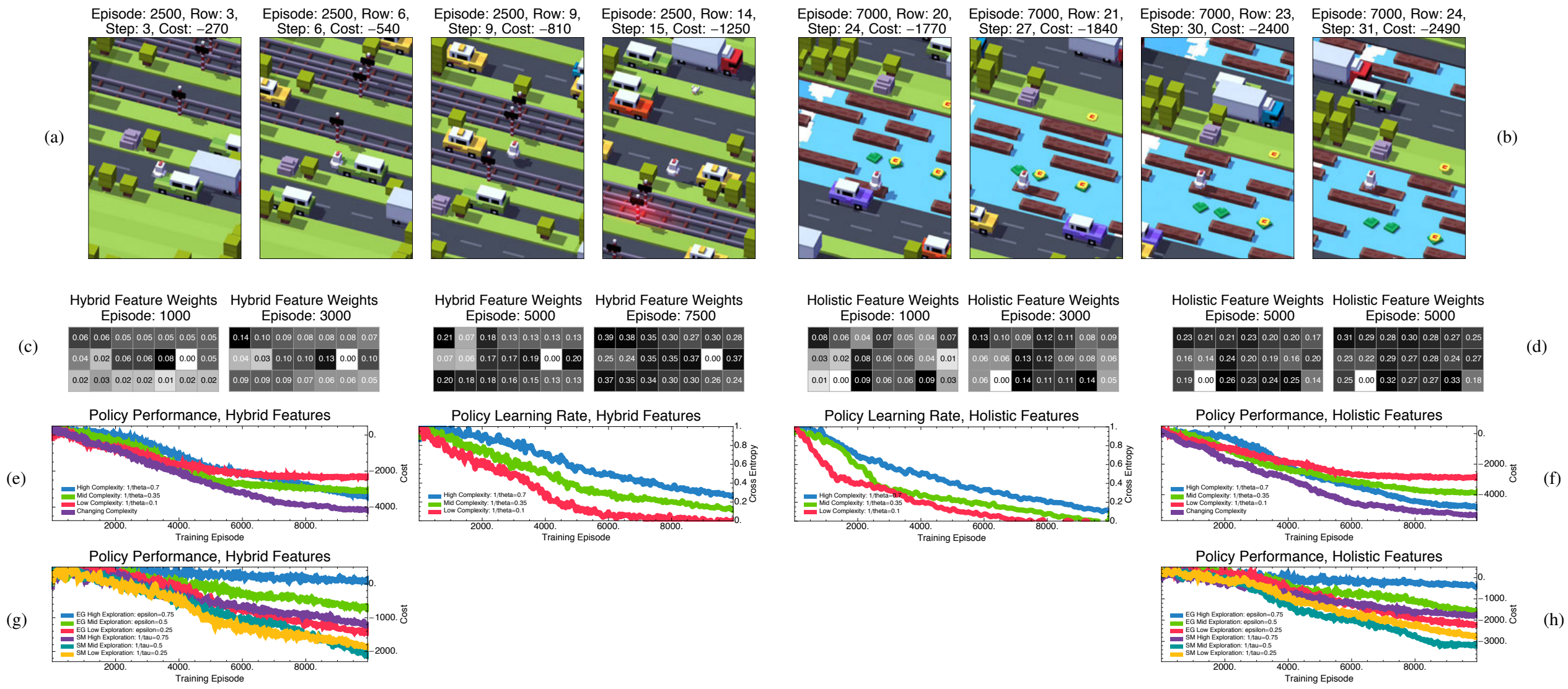

Figure 5.1: Results of the agent's performance during training for the game Crossy Road. In rows (a) and (b) are snapshots of the agent's gameplay behaviors. Early during training, as shown in (a), the agent performs many unnecessary actions in an attempt to understand the environment dynamics. It will often take random actions and get hit by cars frequently. As the training progresses, the agent better understands the relationship between states and actions. As shown in (b), the agent is able to make it far into a given level. It can dodge cars and navigate across waterways. This latter behavior requires the implementation of a multi-modal gameplay strategy, which is not trivial to implement and requires a thorough investigation of the state-action space.

Quantitative results of the learning process are provided in (c)–(h). In rows (c) and (d) are plots of the feature weights for the hybrid and holistic cases, respectively. Darker colors, and hence higher weights, indicate that a given feature is important for the decision-making process. The feature ordering is from left-to-right and top-to-bottom in the same order that they were presented in the discussions. In rows (e) and (f) we plot the policy costs and learning rate for different values of $\theta_t$ for the hybrid and holistic feature cases, respectively. Each colored curve corresponds to a unique test case. We also consider the adaptation of $\theta_t$ according to our cross entropy heuristic. Lower costs indicate a better-performing policy. Lower cross-entropy values indicate fewer changes being made to the policy. The best costs are obtained when $\theta_t$ is allowed to change across each episode versus remaining fixed. Lastly, in rows (g) and (h) are the policy costs produced when using epsilon-greedy and soft-max exploration for the hybrid and holistic feature cases, respectively. Neither of these methods approaches the performance of the uncertainty-based value of information, regardless of the chosen parameters.

The rate at which the cost decreases corresponds to two factors. The first is the feature representation. Contrasting the left-hand plot in figures 5.1(e) with the right-hand plot in figure 5.1(f), it can be seen that the performance of policies for the hybrid feature case lags behind that for the holistic feature case.

The second factor is the policy complexity. When going from mid-complexity policies to those with high complexity, the rate of decrease is lower and the early agent performance is worse. When going from moderately complex policies to those with lower complexity, the situation is reversed. The rate of cost decrease is higher early in the learning process. This suggests a connection between policy complexity and the implementation of agent behaviors.

This connection is partly captured by the policy cross-entropy plots in figures 5.1(e) and 5.1(f). For both feature sets, the cross-entropy is initially close to one, which suggests a great many changes are being made to the policies. The acquisition of new behaviors is prevalent. As the number of episodes grows, the cross-entropy tends toward zero. The rate at which this occurs depends on the policy complexity. Low-complexity policies exhibit the quickest decay of cross-entropy, followed by those that encode longer action sequences. Such results imply that fewer changes are being made to the policies across episodes. Existing behaviors are hence retained.

**Fixed Hyperparameter Effects Discussions.** The cost-decrease change for the different values of $\theta_t$ was expected after the results obtained for the Centipede game. For the highest value of $\theta_t$, the policy space is being finely quantized and searched. The agent is therefore frequently attempting actions that diverge from the current policy. This helps it to uncover a low-cost policy. For lower values of $\theta_t$, the space is coarsely quantized. The exploitation of learned behaviors dominates during learning.

From the supplemental results, it can be seen that deviating from the policy impacts some of the fundamental gameplay behaviors. Initially, the agent behaviors were poor for each of the chosen policy complexities, as shown in figure 5.1(a). Later during training, the distance between the agent and dynamic obstacles on a given row of tiles increased for higher-complexity policies. The amount of time spent waiting on grassy tiles for dynamical obstacles to pass decreased, while the time spent waiting on roadway or railway tiles rose. Each of these changes heightened the chance that the agent would not progress far into the environment. Given enough episodes, policies associated with higher values of $\theta_t$ could perform just as well as, if not better than, those that came about from lower values of $\theta_t$. This claim is corroborated by the supplemental results. It can also be seen by relating the costs at different episodes in figures 5.1(e) and 5.1(f).

For lower values of $\theta_t$, the agent was, initially, better able to cope with the challenge of dodging dynamic obstacles. The agent spent more iterations on grassy tiles planning its future actions. It additionally stayed further away from dynamic obstacles. This is evidenced by the supplemental results. However, when relying on low-complexity policies, the agent may only possess the most basic behaviors for reducing the costs. Rarely were these abilities honed so that the agent could consistently reach tens of rows. The agent would also never seek out coins, even when they were only a tile away. This is because the policy space is being coarsely searched. The exploitation of previously learned behaviors will hence take precedence, after a certain point, over the acquisition of new behaviors. For instance, if the process of seeking coins has not been learned early on, then it likely will not in the later stages of learning. Policies with a higher complexity will thus overtake those of lower complexity after enough episodes have elapsed, as evidenced when contrasting the costs at different points in either figure 5.1(e) or 5.1(f).

From the above simulations, the choice of a feature representation also played a major role in the agent behaviors and hence the policy performance. We found that holistic features, whose results are given in figure 5.1(f), largely outperformed hybrid features, whose results were captured in figure 5.1(e). This was largely due to the informativeness of the former. The holistic features were providing global information about the environment. The agent could therefore better plan actions that would allow it to progress further into the level. It could also better dodge incoming dynamic obstacles and navigate across water tiles, as indicated in figure 5.1(b). Moreover, the holistic features enabled the agent to more effectively seek out coins. Each coin is approximately equal to progressing five to seven rows through the environment. The hybrid features permitted only a modest local view of the environment, in comparison. The agent had less forewarning about dynamic obstacles and an ability to react to them. The agent was also unable to seek coins outside of its limited viewing area, which further limited its cost decrease.

More thorough analyses of the agent behaviors are provided in the online supplement. There, we discuss the witnessed behaviors at various stages of the learning process; video links showcasing these behaviors are also given. We additionally analyze the corresponding feature weights [38] given in figures 5.1(c) and 5.1(d) to better understand how such behaviors emerged.

### 5.2.2 Adjusted Hyperparameter Effects

In this subsection, we assess the performance changes when relying on our cross-entropy heuristic to tune both $\theta_t$ and $\beta_t$ throughout the learning process.

In the previous subsection, we found a relationship between cross entropy and cost decreases. This relationship suggests an efficient scheme to perform reinforcement learning when using uncertainty-based value-of-information exploration. A low action-state complexity should be used, in the beginning, to sparsely survey the entire solution space. High-performing regions will be quickly uncovered when doing so. The policy tables will begin to be populated with meaningful entries, and basic gameplay behaviors will emerge. Eventually, the cross-entropy should dwindle and reach a near-steady-state value. Before this occurs, the policy complexity should be gradually increased to promote a more thorough investigation of the policy space. This determination can be done using threshold tests. Existing gameplay strategies can be refined if this change is made, potentially leading to further cost improvements.

Such a scheme is a departure from the current conventions that involve performing high-exploration searches and progressively scaling back to the exploitation of the current policy. However, when following it, we found that high-performing policies could be found in roughly a third the number of episodes required for the fixed-parameter case. We refer to figures 5.1(e) and 5.1(f) for associated cost statistics. For the simulations in figures 5.1(e) and 5.1(f), we initially considered a value of 0.125 for $\theta_t$ and a value of 0.5 for $\beta_t$. When the cross-entropy reached a chosen threshold, the policy complexity $\theta_t$ was increased by 0.035, while $\beta_t$ was increased by 0.075. Learning was performed for another 20 episodes using these new values. After the conclusion of 20 episodes, another cross-entropy test was performed to determine if the policy complexity should be further increased. This iterative update process repeated until either a maximum value of 1.0 was reached for $\theta_t$ and 1.2 was reached for $\beta_t$. The adjustment also stopped if there was no meaningful policy improvements across 100 episodes.

A cross-entropy threshold value of 0.475 was chosen after experimentation. It represents a point at which the policy is not changing greatly and therefore a slowing of the learning process. Using this threshold value permitted the agent to travel much further into the environment, compared to the results in figure 5.3, once training concluded. These results can be inferred from figure A.2(b). Choosing a high-valued cross-entropy threshold means that the policy complexity will change quickly. The agent may not be capable of sufficiently exploring the state-action space before the policy complexity increases again. Costs can be adversely affected, which is highlighted in figure A.2(a). Not adjusting the policy complexity quickly enough can cause learning stagnation as in figure A.2(c).

**Adjusted Hyperparameter Effects Discussions.** As we noted in the previous subsection, the rate of cost reduction depends on the policy complexity. A priori determining good policy complexities is a challenge. We therefore utilized a heuristic for adjusting the policy complexity using cross-entropy. This adjustment is deforming the free energy of the system from the mutual information term in the Lagrangian (at $\theta_t = 0$) to the distortion term (at $\theta_t = \infty$). It is therefore obtaining the global minimizer of the value of information at low values of $\theta_t$ and tracking the evolution of this minimum as $\theta_t$ is increased.

This heuristic is conceptually supported by statistical thermodynamics. It is known that the entropy of an isolated system can only increase. This leads to the conclusion that entropy is maximized at thermal equilibrium. When the system is not isolated, the maximum entropy principle is replaced by the minimum free energy principle. In the context of the value of information, the Lagrangian is the free energy. This principle states that the Lagrangian cannot increase and therefore reaches its minimum value at thermal equilibrium. By annealing the exploration hyperparameter, the system is always kept either at or near equilibrium. At zero temperature ($1/\theta_t \to 0$, where $\theta_t \to \infty$), the system reaches the minimum energy configuration. This corresponds to the situation where we have the best crisp partitioning of the state space and hence the best policy.

Our simulations demonstrated that iteratively adjusting both $\theta_t$ and $\beta_t$ had a significant impact on agent performance compared to the fixed-parameter case. This performance improvement was made possible due to successive quantizations of the policy space. Without the ability to sparsely sample such spaces, it becomes incredibly difficult to encounter high-performing regions early in the learning process. This partly explains why other search mechanisms must traditionally favor large levels of exploration at first. Likewise, the ability to deterministically explore helped in effectively investigating the quantized space.

Through the use of our heuristic, certain agent behaviors emerged more quickly. The agent learned to avoid static and dynamic obstacles earlier. It also effectively navigated across moving logs. Coins were acquired more frequently too. These implemented agent behaviors were inherently connected with a reduction in costs.

In our simulations, the cross-entropy decayed across many episodes. For certain environments, the cross-entropy may steadily increase after a certain number of episodes. This corresponds to the case where we have increased uncertainty about the action choices due to an influx of so-called neg-information. Such neg-information could be due to widespread changes in the environment dynamics; for instance, the Markov chain underlying the Markov decision process may have vastly different transition probabilities compared to earlier in the learning process. To adequately learn these new dynamics, the exploration hyperparameter $\theta_t$ should be reduced until such a time that the cross-entropy begins to decrease again. Increases in $\theta_t$ according to our above heuristic can then resume.

### 5.2.3 Methodological Comparisons

To provide context for the preceding results, we compare against epsilon-greedy exploration and soft-max selection. These are two of the most well-known stochastic search strategies for reinforcement learning [1].

Epsilon-greedy is based upon the principle of taking random actions and random times and greedily choosing the best action in all other instances. Soft-max selection relies on a weighted-random approach to action selection. It weights each of the actions according to their expected costs and stochastically chooses an arm using those weights.

It is well known that epsilon-greedy exploration can converge to optimal values, in certain situations, as the number of episodes grows. For our application, convergence did not occur within ten thousand episodes, as indicated by figures 5.1(g) and 5.1(h). Moreover, the results are worse than for the uncertainty-based value of information by at least half. The curves in these plots correspond to the following values of $\epsilon$: 0.25, 0.5, and 0.75. Such values correspond to taking a random action every fourth iteration, every other iteration, and three times over four iterations, respectively.

As highlighted in figures 5.1(g) and 5.1(h), soft-max selection yielded modest benefits over epsilon-greedy exploration. It did not, however, achieve similar costs as the uncertainty-based value of information. The curves in these plots correspond to the case of adjusting the following values of $\tau^{-1}$: 0.25, 0.5, and 0.75. These parameter magnitudes, respectively, favor taking the best-cost action frequently, balancing exploration and exploitation, and promoting the near-equi-probable selection of actions.

We attempted to improve these results by making a variety of adjustments. We increased the chance of taking a random action, which only hurt performance over the number of episodes that we considered. We also lowered the random-action-selection rate, but that only yielded a marginal improvement. Training over several thousand more episodes did little to enhance the agent's performance. Adjusting the learning rate and discount factor had only a minimal impact too. The only protocol that we found to consistently improve the policy cost was to slowly anneal $\epsilon$ and $\tau^{-1}$ from a high to a low values. Even then, tens of thousands of episodes were needed to reach the same cost levels as uncertainty-based value-of-information exploration.

**Methodological Comparisons Discussions.** These results highlight the utility of our uncertainty-based value of information search mechanism. Regardless of how we tuned various parameters for either epsilon-greedy or soft-max selection, neither approach was able to reach similar costs as our method in the same number of episodes.

There are two reasons for this. First, such heuristics are unable to determine if a particular region of the policy space has been sufficiently visited. They may therefore randomly select actions that will neither visit new regions of the policy space nor lead to improved costs. Secondly, these heuristics must search over the entire policy space, not a quantized variant of it. It may therefore spend an inordinate number of episodes dealing with related groups of states and not make meaningful revisions to the policy.

In terms of gameplay, both exploration heuristics failed to implement crucial behaviors during the specified episodes. The agents were typically stymied by moving logs. Such sections of the environment can appear early in the level, which limits the potential score if the agents are unable to successfully navigate through them. The agents also typically made unnecessary moves. They would sometimes move laterally, with no apparent gameplay benefit, thereby increasing costs. They would also either prematurely jump into roadways or railways or not move quickly enough away from a dynamic obstacle. Lastly, the agents often failed to collect easily reached coins. The only coins that were picked up were those in the direct path of the agent.

## 6 Conclusions

The traditional value-of-information criterion, when applied to Markov decision processes, quantifies the expected reduction in costs that can be achieved for policies with a specified complexity. We have extended the criterion in this paper. This modified version includes an additional term that measures the divergence between the expected state transition and the actual state transition. It therefore seeks the largest reduction in costs and transition-model uncertainty for under a certain policy complexity. Equivalently, this modified criterion quantifies the simplest policy complexity that produces costs and transition uncertainties below given thresholds.

Optimization of either form of this uncertainty-based value-of-information criterion yields a soft-max-like action-selection process. This process implements a type of policy-space exploration that is stochastically guided. It is stochastic in the sense that actions are selected in a weighted-random fashion according to expected costs and additional constraints. Random walks therefore occur in local neighborhoods of the policy space. It is guided since minimizing the state-transition uncertainty drives the policy iterates to unvisited regions of the policy space.

Such an exploration strategy efficiently investigates the policy space. This functionality stems from the state-space quantization imposed by the expected reward or policy complexity constraints from the value-of-information criterion. Quantization of this space causes related states in local neighborhoods to be grouped and assigned the same action response. By including a state-transition uncertainty minimization component with this criterion, we have provided a means of visiting each of these state groups in a near-uniform manner. As the quantization level increases, the number of state groups decreases. It becomes possible to explore the entire policy space in only a few episodes, at the expense of policy performance. With a large number of state groups, many episodes may be required to exhaustively search the space, yet the policy performance can be high. For some application-dependent quantization amount, there will be a balance between the obtainable performance and expected computational demands.

Our simulations focused on the arcade game Crossy Road. We demonstrated the impact of the quantization amount on the agent behaviors and the corresponding policy costs. We showed that the incorporation of uncertainty-driven search in the value-of-information framework was crucial for obtaining low costs in only a few episodes, regardless of the quantization amount. We additionally highlighted, via cross-entropy, that this reduction in costs was due to an effective learning of new behaviors.

Our simulations suggest a reversal of the conventional learning strategy, at least for our uncertainty-based value-of-information approach. Instead of exploring early in the learning process and exploiting in the later stages, it is advantageous to do the opposite. This led to improved policy performance. Cross-entropy can be used to monitor the learning rate and determine when the state-space quantization should be refined.

Comparatively, other exploration approaches lagged behind in terms of costs. Value-of-information-based policies, for instance, required many more episodes to reach a specified cost. This is because the exploration is purely stochastically driven for this criterion, making it difficult to guarantee that large portions of the policy space will be visited for a finite number of episodes. Similar claims can be made for search heuristics such as epsilon-greedy and soft-max exploration. We also obtained better-performing policies more quickly from uncertainty-based value of information than divergence-to-go. Divergence-to-go does not have the ability to quantize the search space, unlike the value of information. A great many episodes may therefore be wasted exploring highly similar states.

A drawback to the uncertainty-based value-of-information criterion is the use of kernel density estimation for finding the transition-probability distributions. Kernel density estimation is difficult when dealing with many state features, due to the curse of dimensionality. As more features are utilized to describe the states, observations of the state transitions become more sparse and the density estimators take longer to converge to the actual distributions. Our endeavors will focus on two ways to mitigate this issue. For the first approach, we will consider non-linear transformations of the state features to lower-dimensional domains. We will investigate definitions of the divergence term that rely on inner products between state feature vectors in the second approach. The second approach should be more promising, as no information about the states is discarded.